\RequirePackage{amsthm}
\documentclass{sn-jnl}
\usepackage{microtype}
\usepackage{revdiff}
\usepackage{tikz}
\usetikzlibrary{arrows,positioning}
\usepackage{amsmath}
\usepackage{amssymb}
\usepackage{manyfoot}
\usepackage[square,numbers]{natbib}
\newtheoremstyle{thm}
{6pt}
{3pt}
{\itshape}
{}
{\bfseries}
{.}
{.5em}
{}%
\newtheoremstyle{def}
{6pt}
{6pt}
{}
{}
{\bfseries}
{.}
{.5em}
{}%

\theoremstyle{thm}
\newtheorem{theorem}{Theorem}
\newtheorem{lemma}{Lemma}
\newtheorem{corollary}{Corollary}
\newtheorem{definition}{Definition}
\theoremstyle{def}
\newtheorem{example}{Example}

\usepackage{etoolbox}
\AfterEndPreamble{
\setlength\abovedisplayskip{9pt plus 2pt minus 1pt}%
\setlength\belowdisplayskip{9pt plus 2pt minus 1pt}%
\setlength\abovedisplayshortskip{3pt plus 2pt minus 1pt}%
\setlength\belowdisplayshortskip{9pt plus 2pt minus 1pt}%
}

\newcommand{\obl}{\mathop{\mathsf{obl}}}
\newcommand{\perm}{\mathop{\mathsf{perm}}\nolimits}
\NewDocumentCommand{\naf}{s}{
    \mathop{\mathsf{not}
    \IfBooleanF{#1}{\;}} 
}
\newcommand{\dop}{\mathop{\mathsf{dop}}}
\newcommand{\non}{\mathnormal{\sim}}
\newcommand{\sub}{\mathop{\mathrm{Sub}}}
\newcommand{\Jargs}{\mathit{JArgs}}
\newcommand{\Args}{\mathit{Args}}
\newcommand{\Rargs}{\mathit{RArgs}}

\begin{document}

\newif\ifblindreview
\blindreviewfalse 

\title{Deontic Argumentation}
\ifblindreview
\author{}
\affil{}
\else
\author[1]{\fnm{Guido} \sur{Governatori}}\email{g.governatori@cqu.edu.au}
\affil[1]{\orgdiv{School of Engineering and Technology}, 
   \orgname{Central Queensland University}, 
   \orgaddress{\city{Rockhampton}, \country{Australia}}}
\author[2]{\fnm{Antonino} \sur{Rotolo}}\email{antonino.rotolo@unibo.it}
\affil[2]{\orgdiv{Alma AI and Department of Legal Studies}, \orgname{University of Bologna}, \orgaddress{\city{Bologna}, \country{Italy}}}
\fi

\abstract{
    We address the issue of defining a semantics for deontic argumentation that supports weak permission.  Some recent results show that grounded semantics do not support weak permission when there is a conflict between two obligations. We provide a definition of Deontic Argumentation Theory that accounts for weak permission, and we recall the result about grounded semantics. Then, we propose a new semantics that supports weak permission.
}


\maketitle

\section{Introduction}
\label{sec:intro}

Deontic argumentation deals with any  argumentative setting in which the main purpose is establishing whether certain actions or states-of-affairs are obligatory, permitted, or deontically qualified in terms of any other normative position. Hence, its relevance covers several important normative domains, such as the social, ethical \cite{Bench-Capon03}, and the legal one \cite{Bongiovanni2011-POSHIL-2}. 

Deontic argumentation can be technically developed in various ways (see Section \ref{sec:related} for an overview of the literature). As done sometimes in the AI\&Law research \cite{PrakkenS15}, we can assume that legal norms are rules having the form $a_1, \dots , a_n \Rightarrow b$ and we can follow the intuition that deontic arguments are built in an argumentation framework using rules of the form $a_1, \dots , a_n \Rightarrow \obl(b)$, such that   
     $\obl (b)$ holds iff $\obl(b)$ is justified with respect to such an argumentation framework (see \cite{GovernatoriRotoloRiveret,RiveretRotoloSartor}).  

The value of deontic argumentation in the law is that it allows for an account of deontic positions in adversarial, or at any rate dialogical, settings where the determination of what is obligatory or permitted is the result of an exchange of arguments among different subjects. Consider, among others, the case of deontic argumentation in legislative deliberation (see \cite{Meyeretal,icail2029}), or in judicial decision-making (see e.g. \cite{icail23,reka}).

Despite the apparent smoothness in extending well-known formal argumentation techniques to cover \rchange{deontics}{deontic reasoning}, deontic argumentation presents a number of specific challenges, several of them being inherited from deontic logic or from its reconstruction in argumentative settings.

One significant challenge that motivates our research concerns grasping a well-known multifaceted deontic concept, the notion of weak permission \cite{alchourron1971normative,alchourron-bulygin:1984,MakinsonT03,jpl:permission,handbook:permission}. In deontic logic, a permission $\perm (a)$ is derived as a \textit{weak permission} if $\obl (\neg a)$ is not derivable. In judicial contexts, for example, the burden of proof related to this adversarial character of deontic reasoning gives specific argumentative significance to obligations and permissions, including weak permissions \cite{Bongiovanni2011-POSHIL-2}. If a plaintiff argues that a specific action is obligatory (e.g., a duty of care), this claim must be rigorously justified within the argumentative framework of the case. In contrast, a defendant might successfully challenge this obligation by demonstrating that the opposite—a failure to act—is only weakly permitted in the case at stake. This asymmetry underscores the importance of a formal system capable of handling the different strengths of arguments and the varied burdens of proof inherent in deontic judgments. 

Formal argumentation systems can provide the tools to address these challenges.  They offer a structured way to represent deontic statements, reason about their implications, and evaluate the validity of arguments.  By explicitly stating premises, rules of inference, and conclusions, they can shed light on the nature of permissions.  

The primary motivation for our research lies in limitations of some popular argumentation semantics as applied to deontic logic, such as grounded and stable semantics. These semantics inadequately handle weak permissions. These methods often prevent justifications for weak permission due to their inherent limitations in conflict resolution, notably when multiple deontic rules apply simultaneously. To address these limitations, we propose an argumentation semantics that effectively incorporates weak permission by allowing a more nuanced evaluation of arguments. Our approach ensures that both obligations and permissions are considered robustly within argumentative frameworks, promoting a balance between legal precision and interpretative flexibility.

The next sections will explore how formal argumentation frameworks can enhance our ability to navigate the complexities of deontic logic in contexts like these. The layout of the paper is as follows. Section \ref{sec:intuitions} addresses the challenges of deontic argumentation, particularly focusing on weak permission and its theoretical nuances. In Section \ref{sec:background}, we introduce the notion of Deontic Argumentation Theory and recall the results of \cite{weakpermission} for weak permission under the grounded semantics. Section \ref{sec:new} presents our new argumentation semantics tailored for weak permission. \rnew{In Section \ref{sec:discussion}, we discuss the implications of our findings and how they contribute to the broader field of deontic argumentation}. Lastly, Section \ref{sec:related} reviews related work in the domain, and Section \ref{sec:conclusion} summarises our findings and contributions.

\section{Deontic Challenges in Argumentation}\label{sec:intuitions}

What challenges does weak permission pose in an argumentative setting? Consider the following reasoning pattern where $\Vdash$ denotes a generic consequence relation, i.e., a reasoning link between a set of premises ($\Gamma$) and a conclusion:
\begin{equation}\label{eq:weak1}
  \tag{WeakPerm}
  \frac{\Gamma \nVdash \obl (\neg a)}{\Gamma \Vdash \perm (a)}  
\end{equation}
If the logical closure of $\Gamma$ under $\Vdash$ does not contain $\obl (\neg a)$, then we can argue that $a$ is permitted in $\Gamma$. This clearly shows that the precise meaning of \eqref{eq:weak1} substantially depends on the nature and properties of $\Vdash$. Consider for example, Standard Deontic Logic (SDL), which is the deontic interpretation of modal logic $\mathbf{KD}$ built on top of classical propositional logic \cite{handbook:deontic-basic}, \rchange{since}{where} $\perm (a) =_{\mathit{def}} \neg\!\obl (\neg a)$. Moreover, we consider the standard semantic notions of truth in a possible world in a Kripke model, $\vDash$. In SDL, for any model $\mathcal{M}$ and any possible world $w$ we have that $\mathcal{M}, w \nvDash \obl (\neg a)$ iff $\mathcal{M}, w \vDash \neg\!\obl(\neg a)$; but by the duality of obligation and permission in SDL, this is equivalent to $\mathcal{M}, w \vDash \perm (a)$. 
Then \eqref{eq:weak1} trivially holds: if a possible world does not entail the prohibition of $a$, then $a$ is permitted in that world\rnew{\footnote{While \eqref{eq:weak1} holds semantically in a possible world, the property does not hold in general in SDL, and we cannot infer $\Gamma\vdash\perm(\neg a)$ from $\Gamma\nvdash\obl(a)$.}.}

This reading of the deontic closure looks very natural if $\Gamma$ represents a criminal legal system, where the closure rule \emph{nullum crimen sine lege} is nothing but the idea that any $a$ which is not explicitly prohibited (by criminal law) is permitted (by criminal law) \cite{alchourron1971normative,alchourron-bulygin:1984,icail23}.

Suppose now that
\begin{equation}
\label{eq:conflict} \obl (a), \obl (\neg a)\in \Gamma.
\end{equation}
Clearly, this case becomes trivial in SDL and leads to a deontic explosion \cite{handbook:deontic-basic}, since it is based on classical logic: despite the fact that \eqref{eq:weak1} cannot be used as such to license the conclusion of $\perm (a)$ or $\perm (\neg a)$, we obtain both of them either because $\obl (a)$ implies in SDL $\neg\!\obl (\neg a)$ (resp. $\obl (\neg a)$ implies in SDL $\neg\!\obl (a$)), or because $\Gamma$ is inconsistent and supports the conclusion of any \rchange{deontic statement}{statement (deontic or not)}. 

Assume now to abandon classical logic and SDL in favour of any logical system which tolerates normative conflicts. 
If we reconstruct \eqref{eq:weak1} in defeasible reasoning by using a suitable argumentation framework---and assuming as usual that a conclusion of a justified argument is a justified conclusion---
then the following is an informal reformulation of \eqref{eq:weak1}:
\begin{align}
  \frac{\obl (\neg a) \text{ is not justified in }\Gamma}{\perm (a) \text{ is justified in } \Gamma}  \label{eq:weak2}
  \tag{A-WeakPerm} & \qquad 
  \frac{\perm (a) \text{ is justified in }\Gamma}{\obl (\neg a) \text{ is not justified in } \Gamma} 
\end{align}
This argumentative reading of weak permission paves the way for interesting deontic inquiries. \rchange{
Again, suppose that (1) holds:}{Suppose now, we have two applicable norms, one that makes $a$ obligatory and the second that forbids it and there is no mechanism to determine whether one of the two norms prevails over the other
} this intuitively means that we have two arguments supporting both $\obl (a)$ and $\obl (\neg a)$ in our argumentation framework. 
\rchange{It seems clear that that \emph{the nature of weak permission depends on the argumentation semantics adopted.}}{ 
The issue whether we can conclude that $a$ or $\neg a$ are weakly permitted \emph{depends on the argumentation semantics adopted}.}
To better explain this point, consider the following example.

\begin{example}\label{ex:ambiguity1}
Consider the following deontic scenario:
\begin{align*}
n_1: & \; f_1 , \dots , f_n \Rightarrow \obl (a) \\
n_2: & \; g_1 , \dots , g_m \Rightarrow \obl (\neg a) \\
n_3: & \; \perm (a), \perm (\neg a) \Rightarrow \obl (c)\\
n_4: & \; h_1, \dots , h_k \Rightarrow \obl (\neg c)
\end{align*}
Norm $n_1$ prescribes $\obl (a)$ under condition $f_1 , \dots , f_n$. Concurrently, the second norm $n_2$, independent of the first, forbids $a$ when $g_1, \dots ,  g_m$ hold. Both norms are at the same hierarchical level in the normative system, meaning there are no  criteria to determine if one norm overrides the other. Norm  $n_3$ requires $\obl (c)$ if $a$ is legally facultative, i.e., if $\perm (a)$ and $\perm (\neg a)$ are the case \cite{Sartor:2005}. 
Finally, norm $n_4$ prescribes $\obl (\neg c)$ under conditions $h_1, \dots , h_k$. However, $n_4$ is weaker that $n_3$, meaning that $n_3$ overrides $n_4$ if both are applicable.
\end{example}
Norms $n_1$ and $n_2$ \rchange{bring to}{potentially yield} a genuine deontic conflict; specifically, the deontic conclusion that “$a$ is obligatory” is ambiguous due to two conflicting reasoning chains, one supporting this conclusion and another opposing it. 

\begin{figure}[h]
\centering
\begin{tikzpicture}[
    node distance=2.5cm,
    argument/.style={circle, draw, minimum size=1.5cm, thick},
    attack/.style={->, thick} 
]
    \node[argument] (A) at (0,0) {$\obl(a)$};
    \node[argument] (B) at (0,3) {$\obl(\neg a)$};
    \node[argument] (C) at (2.5,1.5) {$\obl(c)$};
    \node[argument] (D) at (5,1.5) {$\obl(\neg c)$};
    
    \draw[attack] (A) to[bend left=20] (B);
    \draw[attack] (B) to[bend left=20] (A);
    \draw[attack] (A) to (C);
    \draw[attack] (B) to (C);
    \draw[attack] (C) to (D);
\end{tikzpicture}
\caption{\rnew{Argument Graph for Weak Permission and Facultative Conclusions where nodes denote arguments (with their conclusion) and arrows denote attacks.}}
\label{fig:floating}
\end{figure}

\rchange{If we import \emph{sic et simpliciter} the standard analysis 
of the example under the grounded semantics \cite{Dung1995}\footnote{See Section~\ref{sec:background} for the technical details of grounded semantics.} 
we obtain the argumentation structure depicted in Figure~\ref{fig:floating},assuming the conditions of the rules apply and that arguments are identified by their conclusions. The arguments for $n_1$ and $n_2$ attack each other (given that one rule establishes the obligation of $a$ whereas the other forbids it).  For the arguments for $\obl(c)$ we obtain an argument from $n_1$ and $n_2$ (using the obligation implies permission of standard deontic logic). This argument attacks the argument for $n_4$, since we assumed that $n_3$ defeats $n_4$. However, $n_1$ and $n_2$ attack (or better undermine) the argument for $n_3$, given the deontic incompatibility between $\obl(p)$ and $\perm(\neg p)$. Under the standard interpretation of grounded semantics, the only justified argument is the argument for $\obl(\neg c)$, which is the only justified conclusion. }{
If we import, \emph{sic et simpliciter}, the standard analysis of the example under grounded semantics \cite{Dung1995}\footnote{See Section~\ref{sec:background} for the technical details of grounded semantics.},
we obtain the argumentation structure depicted in Figure~\ref{fig:floating},assuming the conditions of the rules apply and that arguments are identified by their conclusions. The arguments for $n_1$ and $n_2$ attack each other (given 
For $\obl(c)$, we obtain an argument from $n_1$ and $n_2$, using the principle from standard deontic logic that obligation implies permission. This argument attacks the argument based on $n_4$, since we assumed that $n_3$ defeats $n_4$. However, $n_1$ and $n_2$ also attack, or more precisely undermine, the argument based on $n_3$, given the deontic incompatibility between $\obl(p)$ and $\perm(\neg p)$. Under the standard interpretation of grounded semantics, the only justified argument is the one for $\obl(\neg c)$, which is therefore the only justified conclusion.}

Under the strong permission reading of $\perm(a)$ and $\perm(\neg a)$ in $n_3$ it is not possible to obtain $\obl(c)$ when at least one of $n_1$ and $n_2$ is applicable.  Let us consider applying \eqref{eq:weak2}. As we have just seen, $\obl(a)$, $\obl(\neg a)$, and $\obl(c)$ are not justified. Thus, using \eqref{eq:weak2} we obtain that $\perm(a)$, $\perm(\neg a)$, and $\perm(\neg c)$ are justified (as weak permissions). Given that we have $\perm(a)$ and $\perm(\neg a)$, we should be able to conclude $\obl(c)$ using $n_3$, and then $n_4$ would no longer be justified. However, if we adopt \eqref{eq:weak2} on top of the grounded semantics, we still obtain that $\obl(\neg c)$ is  a justified conclusion.

In general, Example \ref{ex:ambiguity1} offers a scenario very  familiar in deontic reasoning. However, a deeper understanding of it must require a broader exploration of how this type of deontic conflict can impact subsequent deontic reasoning chains.  

\rnew{
Let us ground the idea above in a concrete legal scenario. Article 520 of the Italian Code of Criminal Procedure establishes that a judge must issue a sentence of acquittal if (i) the fact does not constitute a crime or (ii) the fact is not classified by law as a crime.
Moreover, the same article states that the judge must issue a sentence of acquittal when there is not enough evidence to conclude that (a) the fact constitutes a crime, (b) the defendant committed the crime, or (c) the fact is classified by law as a crime. Here we can identify two main patterns.

The first pattern is that there is an obligation (or prohibition) in force, but there is not enough evidence to conclude that the defendant committed the crime, so the content of the obligation is not satisfied.

The second pattern is that there is not enough evidence to conclude that an obligation (or prohibtion) is in force and applicable to the case. Thus, if there is no obligation, the fact is (weakly) permitted.

The first part of the article can be represented by the following norms:
\begin{align*}
    n_1\colon & \obl(f), \neg f \Rightarrow \obl(\mathit{Acquittal})\\
    n_2\colon & \perm(f), f \Rightarrow \obl(\mathit{Acquittal})
\end{align*}
The second part is delegated to reasoning with the facts of the case and the norms of the underlying legal system in order to establish whether the obligation is in force, whether the fact corresponds to the content of the obligation, and whether there is enough evidence to conclude that the obligation is in force and applicable to the case.

Consider now the infamous Sea Watch 3 case. In 2019, the Sea Watch 3, an NGO ship, rescued 53 migrants in distress in the Mediterranean Sea. The Italian Government issued a decree prohibiting the ship from rescuing migrants in the waters of the Italian contiguous zone. At the same time, however, maritime law established that any ship in proximity to a vessel in distress has an obligation to rescue people at sea. Accordingly, we can represent the case by the following rules:
\begin{gather*}
    r_1\colon\mathit{distress}, \mathit{proximity} \Rightarrow \obl(\mathit{rescue})\\
    r_2\colon\mathit{SeaWatch}, \mathit{migrants}, \mathit{ItalianContiguousZone} \Rightarrow \obl(\neg\mathit{rescue})
\end{gather*}  
In addition, the two norms in Article 520 can be instantiated as follows:
\begin{gather*}
    n_1\colon \obl(\mathit{rescue}), \neg\mathit{rescue} \Rightarrow \obl(\mathit{Acquittal})\\
    n_2\colon \perm(\mathit{rescue}), \mathit{rescue} \Rightarrow \obl(\mathit{Acquittal})
\end{gather*}
Given that the conditions in the antecedents of rules $r_1$ and $r_2$ are satisfied, we have a conflict between $\obl(\mathit{rescue})$ and $\obl(\neg\mathit{rescue})$. The issue, then, is whether we can conclude $\perm(\mathit{rescue})$ and $\perm(\neg\mathit{rescue})$ as weak permissions and, on that basis, derive $\obl(\mathit{Acquittal})$ using $n_2$, or whether rescuing the migrants was not permitted at all.}\footnote{\rnew{In response to the Sea Watch 3 incident, an Italian tribunal held that international maritime law prevailed over the Italian Government decree in that specific case. However, scholars of international maritime law debated the proper course of action.}}

\section{Grounded Semantics and Weak Permission}
\label{sec:background}
In this section, we recall the results of \cite{weakpermission} that show that grounded semantics does not support weak permission when there is a conflict between two obligations. Moreover, the section sets the definition for the notion of a deontic argumentation theory. The definitions of argument and attacks are common to the grounded semantics and the semantics we are going to develop in the next section. 

The language of a deontic argument is built from a set of literals ($\mathrm{Lit}$), where a literal is either an atom ($l$) or its negation ($\neg l$). In addition, we extend the language with \emph{deontic literals}. The set of deontic literals is defined as
\(
\{\dop(l) | l\in \mathrm{Lit} \}
\)
where $\dop$ is a deontic operator, more precisely $\dop\in\{\obl,\perm,\perm_w\}$ (for obligation, (strong) permission and weak permission respectively). Given a literal $l$, we use $\non l$ to denote the complement of $l$; more precisely, if $l$ is an atomic proposition, then $\non l=\neg l$. If $l$ is a negated atomic proposition $l=\neg m$, then $\non l=m$.
Notice that we do not allow the negation outside the deontic operator, we assume that negation in front is pushed inside the deontic operator using the following rules: $\neg\!\obl(l)\mapsto\perm(\non l)$, $\neg\!\perm(l)\mapsto\obl(\non l)$, and $\neg\!\perm_w(l)\mapsto\obl(\non l)$. Essentially, we use the duality between the operators to move the negation in the scope of a deontic operator.

Arguments are built from rules, where a rule has the following format:
\[
a_1,\dots a_n \Rightarrow c
\]
where $\{a_1,\dots,a_n\}$ is a (possibly empty) set of literals and deontic literals, and $c$ is either a literal or deontic literal but not a weak permission (i.e., $c\neq\perm_w(l)$, otherwise it would be an explicit permission).

\begin{definition}
\label{def:theory}
A \emph{Deontic Argumentation Theory} is a structure 
\[(F,R)\]
where $F$ is a (finite and possibly empty) set of literals (the fact or assumption of the theory), and $R$ is a (finite) set of rules. 
\end{definition}
The key concept of an argumentation theory is the notion of an argument. Definition~\ref{def:argument} below defines what an argument is. Each argument $A$ has associated with it, its conclusion $C(A)$ and its set of sub-arguments $\sub(A)$.
\begin{definition}
\label{def:argument}
Given a Deontic Argumentation Theory $(F,R)$,  $A$ is an \emph{argument} if $A$ has one of the following forms: 
\begin{enumerate}
    \item \label{it:weak} $A=\perm_w(l)$ for any literal \rchange{$l\in\mathcal{L}$}{$l\in\mathrm{Lit}$}, the conclusion of the argument $C(A)=\perm_w(l)$, and $\sub(A)=\{A\}$.
    \item \label{it:fact} $A=a$ for $a\in F$; $C(A)=a$ and $\sub(A)=\{A\}$. 
    \item \label{it:normal} $A=A_1,\dots,A_n\Rightarrow c$, if there is a rule $a_1,\dots,a_n\Rightarrow c$ in $R$ such that for all $a_i\in\{a_1,\dots, a_n\}$
    there is an argument $A_i$ such that $C(A_i)=a_i$; $C(A)=c$ and $\sub(A)= \{A\}\cup\sub(A_1)\cup\dots\cup\sub(A_n)$.
    \item \label{it:strong} $A=B \Rightarrow \perm(l)$, if $B$ is an argument such that $C(B)=\obl(l)$; $C(A)=\perm(l)$, and $\sub(A)=\{A\}\cup\sub(B)$.
\end{enumerate}    
We call the weak permission arguments (arguments defined by clause \eqref{it:weak} above) \emph{imaginary arguments}. Arguments defined by the other clauses are called \emph{natural arguments}.
The set of all arguments for a Deontic Argumentation Theory is $\Args$.
\end{definition} 
\noindent
Condition \eqref{it:weak} encodes the idea that weak permission is the failure to derive an obligation to the contrary (more on this \rold{point is offered} when we discuss the notion of attack between arguments). Thus, by default, every literal is potentially weakly permitted, and we form an argument for this type of conclusion. 
Condition \eqref{it:fact} gives the simplest form of an argument. We have an argument for $a$ if $a$ is one of the assumptions/facts of a case/theory. 
Condition \eqref{it:normal} allows us to form arguments by forward chaining rules. Thus, we can form an argument from a rule, if we have arguments for all the elements of the body of the rule. The way the condition is written allows us to create arguments from rules with an empty body.  Finally, condition \eqref{it:strong} corresponds to the D axiom of Standard Deontic Logic.

\begin{definition}[Attack]
\label{def:attack}
  Let $A$ and $B$ be arguments. $A$ \emph{attacks} $B$ ($A > B$) iff
  \begin{enumerate}
    \item \label{attack:weak} $B=\perm_w(l)$ and $C(A)=\obl(\non l)$;
    \item \label{attack:neg} $\exists B'\in\sub(B)$, $C(B')=l$ and $C(A)=\non l$;
    \item \label{attack:obl} $\exists B'\in\sub(B)$, $C(B')\in\{\obl(l),\perm(l)\}$ and $C(A)=\obl(\non l)$;
    \item \label{attack:perm} $\exists B'\in\sub(B)$, $C(B')=\obl(l)$, and $C(A)=\perm(\non l)$. 
  \end{enumerate}
\end{definition}
As we alluded to \rold{above and as we have seen} in Section~\ref{sec:intro}, the idea of weak permission is the negation as failure of the obligation to the contrary. Thus, in Definition~\ref{def:argument}, we create an argument for the weak permission for any literal $l$; however, this argument is attacked by any argument for $\obl(\non l)$ (condition \eqref{attack:weak} above and notice this is the only case where the attack is not symmetrical). The rest of the conditions define an attack when the two arguments have opposite conclusions: condition \eqref{attack:neg} covers the case of plain literals, while conditions \eqref{attack:obl} and \eqref{attack:perm} are reserved for deontic literals. Specifically, we have opposite deontic conclusions when one of the two is an obligation for a literal, and the other is either an obligation or a permission for the opposite literal. Furthermore, an argument attacks another argument when the conflict is on the conclusion of the second argument (this corresponds to the notion of rebuttal) or when there is a conflict with one of the sub-arguments of the attacked argument (known as undercutting attack).

\begin{example}
    \label{ex:extensive}
    Let us consider the Deontic Argumentation Theory $(F,R)$ where $F=\{a\}$ and $R$ contains the following rules:
    \[\begin{array}{c@{\qquad}c@{\qquad}c}
        r_1\colon \Rightarrow\obl(p) &
        r_3\colon \obl(p) \Rightarrow \obl(q) &
        r_5\colon \rnew{a,} \perm_w(p) \Rightarrow s\\
        r_2\colon \Rightarrow\obl(\neg p) &
        r_4\colon \Rightarrow\obl(\neg q) & 
        r_6\colon \perm(q) \Rightarrow t
    \end{array}\]
    \allowdisplaybreaks
    The set of natural arguments contains $A_0\colon a$ (given that $a$ is a fact) and the following arguments built from the above rules:
    \[\begin{array}{c@{\qquad}c@{\qquad}c@{\qquad}c}
    A_1\colon \Rightarrow \obl(p) &
    A_2\colon \Rightarrow \obl(\neg p) &
    A_3\colon A_1 \Rightarrow \obl(q) \\
    A_4\colon \Rightarrow \obl(\neg q) &
    A_5\colon A_1, \Rightarrow \perm(p) &
    A_6\colon A_2 \Rightarrow \perm(\neg p) \\
    A_7\colon A_3 \Rightarrow \perm(q) &
    A_8\colon A_4 \Rightarrow \perm(\neg q) &
    A_9\colon A_0, I_1 \Rightarrow s\\
    &
    A_{10}\colon A_7 \Rightarrow t
    \end{array}\]
Moreover, for each \rchange{propositional letter}{atomic proposition} $a,p,q,s,t$ we have two imaginary arguments, one for the \rchange{letter}{proposition} and one for its negation. Specifically, we consider the following imaginary arguments:
\begin{align*}
    I_1\colon \perm_w(p) &&
    I_2\colon \perm_w(\neg p) &&
    I_3\colon \perm_w(q) &&
    I_4\colon \perm_w(\neg q)
\end{align*}
Arguments \rchange{$A_0, A_1, A_3$}{$A_0, A_1, A_2$} and $A_4$ do not have proper subarguments, and have only themselves as their subarguments.
\begin{gather*} 
\sub(A_3) = \{A_1,A_3\} \qquad
\sub(A_5) = \{A_1,A_5\} \qquad
\sub(A_6) = \{A_2,A_6\} \\
\sub(A_7) = \{A_1,A_3,A_7\} \qquad
\sub(A_8) = \{A_4,A_8\} \\
\sub(A_9) = \{A_0,I_1,A_9\}\qquad
\sub(A_{10}) = \{A_1,A_3,A_7,A_{10}\}
\end{gather*}
The attack relation is as follows (for each argument $A$, we list the arguments that $A$ attacks):
\begin{equation*}
\begin{aligned}
 A_1&\colon A_2, A_6, I_2 &&\quad&
 A_2&\colon A_1, A_3, A_5, I_1, A_{10}\\
 A_3&\colon A_4, A_8, I_4 &&\quad&
 A_4&\colon A_3, A_7, I_3, A_{10}\\
 A_5&\colon A_2, A_6 &&\quad&
 A_6&\colon A_1, A_3, A_5, A_7, A_{10}\\
 A_7&\colon A_4, A_8 &&\quad&
 A_8&\colon A_3, A_7, A_{10}.
\end{aligned}
\end{equation*}
\end{example}
We are ready to recall the standard definitions of Dung \rnew{argumentation}semantics \cite{Dung1995}\rold{ for argumentation theories}.
\begin{definition}[Dung Semantics]
\label{def:semantiche}
    Let $(F,R)$ be a Deontic Argumentation Theory, and $S$ be a set of arguments. Then:
\begin{itemize}
\item $S$ is conflict free iff $\forall X, Y \in S: X \not> Y$.
\item $X \in \Args$ is acceptable with respect to $S$ iff $\forall Y \in\Args$ such that $Y>X$ $: \exists Z \in S$ such that $Z>Y$.
\item $S$ is an admissible set iff $S$ is conflict free and $X \in S$ implies $X$ is acceptable w.r.t. $S$.
\item $S$ is a complete extension iff $S$ is admissible and if $X \in\Args$ is acceptable w.r.t. $S$ then $X \in S$.
\item $S$ is the grounded extension iff $S$ is the set inclusion minimal complete extension.
\item $S$ is a stable extension iff $S$ is conflict free and $\forall Y\notin S$, $\exists X\in S$ such that $X>Y$.
\end{itemize}
\end{definition}

\begin{definition}[Justified Argument]
\label{def:justified-a}
    Let $D=(F,R)$ be a Deontic Argumentation Theory, an argument $A$ is \emph{sceptically justified} under a semantic $T$ iff \rnew{1) an extension for $D$ exists and 2)} $A\in S$ for all sets of arguments $S$ that are an extension under $T$.
\end{definition}

\begin{definition}[Justified Conclusion]
\label{def:justified-c}
    A literal or a deontic literal \rchange{$l\in\mathcal{L}$}{$l\in\mathrm{Lit}$} is a \emph{Justified conclusion} under a semantics $T$ iff \rnew{1) an extension for $D$ exists and 2)} for every extension $S$ under $T$, there is an argument $A$ such that $A\in S$ and $C(A)=l$.
\end{definition}%
\rnew{Notice that, in the definitions above, we require that an extension exist for the theory. This accounts for cases in which the semantics does not guarantee the existence of an extension, such as the stable semantics. In such cases, we do not want to conclude (vacuously) that any argument or conclusion is justified.}

\begin{example}
\label{ex:basic}
Let us consider a Deontic Argumentation Theory where $F=\emptyset$ and $R$  contains the two rules 
\begin{align*}
    r_1\colon {} \Rightarrow \obl(a) &&
    r_2\colon {} \Rightarrow \obl(\neg a)
\end{align*}
This theory has the following arguments:
\begin{align*}
A_1&: \perm_w(a) &
A_3&: {}\Rightarrow \obl(a) &
A_5&: A_3 \Rightarrow \perm(a)\\
A_2&: \perm_w(\neg a) &
A_4&: {}\Rightarrow \obl(\neg a) &
A_6&: A_4 \Rightarrow \perm(\neg a)
\end{align*}
For the attack relation, we have the following instances
\begin{align*}
    A_3 > A_2 && 
    A_3 > A_4 &&
    A_3 > A_6 &&
    A_5 > A_4 &&
    A_5 > A_6 \\
    A_4 > A_1 &&
    A_4 > A_3 &&
    A_4 > A_5 &&
    A_6 > A_3 &&
    A_6 > A_5
\end{align*}
\rnew{The following diagram depicts the argumentation graph of the theory (where the arrows represent the attack relation).
\begin{center}
\begin{tikzpicture}[
    node distance=2cm,
    argument/.style={draw, circle, minimum size=1cm, inner sep=0pt},
    attack/.style={->, thick}
]

\node[argument] (A1) {$A_1$};
\node[argument] (A2) [below=of A1] {$A_2$};
\node[argument] (A3) [right=of A1] {$A_3$};
\node[argument] (A4) [right=of A2] {$A_4$};
\node[argument] (A5) [right=of A3] {$A_5$};
\node[argument] (A6) [right=of A4] {$A_6$};

\draw[attack] (A3) to (A2);
\draw[attack] (A3) to (A4);
\draw[attack] (A3) to (A6);
\draw[attack] (A5) to (A4);
\draw[attack] (A4) to (A1);
\draw[attack] (A4) to (A3);
\draw[attack] (A4) to (A5);
\draw[attack] (A6) to (A3);

\end{tikzpicture}
\end{center}}
It is easy to verify that $\{\}$ is a complete extension (and trivially, it is the minimal complete extension w.r.t.\ set inclusion). Thus, it is the grounded extension of the theory. Accordingly, there is no argument in the grounded extension such that its conclusion is either $\perm_w(a)$ or $\perm_w(\neg a)$. Hence, $\perm_w(a)$ and $\perm_w(\neg a)$ are not justified conclusions under \rnew{the} grounded semantics. 

When we consider the stable semantics, we have the following two extensions:
\begin{align*}
    \{A_1, A_3, A_5\} &&
    \{A_2, A_4, A_6\}
\end{align*}
Clearly, $\perm_w(a)$ is a conclusion of the first extension but not of the second one; conversely, $\perm_w(\neg a)$ is a conclusion of the second extension but not of the first one. Consequently, $\perm_w(a)$ and $\perm_w(\neg a)$ are not justified conclusions under \rnew{the} stable semantics.
\end{example}
The above example should suffice to show that weak permission is not supported by grounded and stable semantics when a deontic conflict exists: we have a scenario where we fail to conclude that an obligation, but at the same time, we cannot conclude the weak permission of the opposite.  However, we can generalise the result; the result holds for any theory with a conflict between two applicable obligation rules. This is formalised by the following definition. 

\begin{definition}
\label{def:conflictual}
    A Deontic Argumentation Theory is \emph{conflictual} when it contains a pair of rules $b_1,\dots,b_n\Rightarrow\obl(c)$ and $d_1,\dots d_m \Rightarrow \obl(\neg c)$,
    such that there are arguments $B_i$ with conclusion $b_i$, $1\leq i\leq n$, and $D_j$ with conclusion $d_j$, $1\leq j\leq m$. We will call $c$ the \emph{conflicted} literal.
\end{definition}

\begin{theorem}
\label{thm:impossibility}
   Let $D=(F,R)$ be a conflictual theory. For any conflicted literal $l$, $\perm_w(l)$, $\perm_w(\neg l)$ are not justified conclusions under grounded and stable semantics.    
\end{theorem}
\begin{proof}
Let $l$ be a conflicted literal. By the definition of argument, the theory $D$ and the fact that the theory is conflicted, we have the following arguments: 
\begin{align*}
    A_1&: \perm_w(l) &    A_3&: B_1,\dots, B_n \Rightarrow\obl(l)\\
    A_2&: \perm_w(\neg l) &
    A_4&: D_1,\dots, D_m \Rightarrow\obl(\neg l)
\end{align*}   
such that $A_3 > A_2$ and $A_4 > A_1$. Given that there are arguments attacking $A_1$ and $A_2$, these two arguments are not in the minimal complete extension. Accordingly, $\perm_w(l)$ and $\perm(\neg l)$ are not justified conclusions under the grounded semantics.  

\rnew{For the stable semantics, if there are no extensions, then there are no justified conclusions. If there are extensions, t}hen
given the attack relationship among $A_1$, $A_2$, $A_3$ and $A_4$, we can conclude that there are at least two extensions $E_1$ and $E_2$ such that $A_1,A_3\in E_1$ and $A_2,A_4\notin E_1$, and $A_2,A_4\in E_2$ and $A_1,A_3\notin E_2$. Hence, there is an extension where no argument has $\perm_w(l)$ as its conclusion, and there is an extension where no argument has $\perm_w(\neg l)$ as its conclusion. Therefore, $\perm_w(l)$ and $\perm_w(\neg l)$ are not justified conclusions under \rnew{the} stable semantics. 
\end{proof}

\rnew{
\begin{example}\label{ex:sea-watch-arg}
Let us consider the Deontic Argumentation Theory modelling the Sea Watch 3 scenario discussed in Section~\ref{sec:background}. The set of facts $F$ and the set of rules $R$ for the case are as follows:
\begin{align*}
F = \{ & \mathit{distress}, \mathit{proximity}, \mathit{SeaWatch}, \mathit{migrants}, \mathit{ItalianContiguousZone}\}\\
R = \{ &
r_1\colon\mathit{distress}, \mathit{proximity} \Rightarrow \obl(\mathit{rescue})\\
 & r_2\colon\mathit{SeaWatch}, \mathit{migrants}, \mathit{ItalianContiguousZone} \Rightarrow \obl(\neg\mathit{rescue})\\
 & n_1\colon \obl(\mathit{rescue}), \neg\mathit{rescue} \Rightarrow \obl(\mathit{Acquittal})\\
 & n_2\colon \perm_w(\mathit{rescue}), \mathit{rescue} \Rightarrow \obl(\mathit{Acquittal})
\}
\end{align*}
From this theory, we can construct the following arguments:
\begin{align*}
    I_1&: \perm_w(\mathit{rescue}) &
    I_2&: \perm_w(\neg\mathit{rescue}) \\
    A_1&: \mathit{distress} &
    A_2&: \mathit{proximity} \\
    A_3&: \mathit{SeaWatch} &
    A_4&: \mathit{migrants} \\
    A_5&: \mathit{ItalianContiguousZone} &
    A_6&: \mathit{rescue} \\
    A_7&: A_1, A_2 \Rightarrow \obl(\mathit{rescue}) &
    A_8&: A_3, A_4, A_5 \Rightarrow \obl(\neg\mathit{rescue}) \\
    A_9&: I_2, A_6 \Rightarrow \obl(\mathit{Acquittal}) 
\end{align*}
The attack relation is as follows:
\begin{align*}
    A_7 > I_2 && A_7 > A_8 && A_7 > A_9 \\
    A_8 > I_1 && A_8 > A_7 
\end{align*} 
Accordingly, we can depict the argumentation graph of the theory as follows (where the arrows represent the attack relation).
\begin{center}
\vspace{0.5cm}    
\begin{tikzpicture}[
    node distance=1cm,
    argument/.style={draw, circle, minimum size=1cm, inner sep=0pt},
    attack/.style={->, thick}]
    \node (a1) [argument] {$A_1$};
    \node (a2) [argument,right=of a1] {$A_2$};
    \node (a3) [argument,right=of a2] {$A_3$};
    \node (a4) [argument,right=of a3] {$A_4$};
    \node (a5) [argument,right=of a4] {$A_5$};
    \node (a6) [argument,right=of a5] {$A_6$};

    \node (a7) [argument, below=of a2] {$A_7$};
    \node (a8) [argument, below=of a5] {$A_8$};
    \node (i2) [argument, below=of a7] {$I_2$};
    \node (i1) [argument, below=of a8] {$I_1$};

    \node(a9) [argument, right=of i2, xshift=1cm] {$A_9$};

    \draw[attack] (a7) to [bend left=10] (a8);
    \draw[attack] (a8) to [bend left=10] (a7);
    \draw[attack] (a7) to (i2);
    \draw[attack] (a8) to (i1);
    \draw[attack] (a7) to (a9);
\end{tikzpicture}
\vspace{0.5cm}
\end{center}    
In this theory, the arguments $A_1, A_2, A_3, A_4, A_5, A_6$ are undisputed facts, but there is a conflict between $A_7$ and $A_8$. The two arguments attack each other, and they also attack the arguments for the weak permissions. This means that, under grounded semantics, the only complete extension is
\[
\{A_1, A_2, A_3, A_4, A_5, A_6\}
\]
which does not include any argument for a weak permission. Hence $\perm_w(\mathit{rescue})$ and $\perm_w(\neg\mathit{rescue})$ are not justified conclusions under the grounded semantics.  

$A_9$ cannot be in the extension because it is attacked by $A_7$ and $A_7$ is not attacked by any argument in the extension. Moreover, $I_1$ and $I_2$ cannot be in the extension because they are attacked by $A_8$ and $A_7$ respectively, and neither of these two arguments is attacked by any argument in the extension. However, we failed to establish that there is enough evidence to conclude $\obl(\neg\mathit{rescue})$; thus according to Article 520 a sentence of acquittal should be issued. However, this is not possible under the grounded semantics, because we cannot conclude $\perm_w(\mathit{rescue})$. 

For the stable semantics, we have the following two extensions:
\begin{align*}
    \{A_1, A_2, A_3, A_4, A_5, A_6, A_8, A_9, I_2\} &&
    \{A_1, A_2, A_3, A_4, A_5, A_6, A_7, I_1\}
\end{align*}
\end{example}
Here we have one extension where  $\obl(\mathit{rescure})$ (and then $\perm_w(\mathit{rescue}))$ hold, another extension in which $\obl(\neg\mathit{rescue})$ holds. Hence, there is ``evidence'' for both obligations. Thus, the situation is similar to the case of the grounded semantics, where we have a conflict between two obligations, and we cannot conclude the weak permission of the opposite.  More specifically, we cannot conclude $\perm_w(\mathit{rescue})$, again in contradiction with Article 520. 
}

\begin{definition}
A literal $l$ is \emph{facultative} if $\perm(l),\perm(\neg l)$ are justified conclusions. $l$ is \emph{weakly facultative} if $\perm_w(l),\perm_w(\neg l)$ are justified conclusions. 
\end{definition}

Based on this definition, we have the following immediate corollary of Theorem~\ref{thm:impossibility}.

\begin{corollary}
    \label{cor:facultative}
    Let $D=(F,R)$ be a conflictual theory. For any conflicted literal $l$, $l$ is not weakly facultative under the grounded semantics and stable semantics.
\end{corollary}

Corollary~\ref{cor:facultative} establishes that, in case of a conflictual literal $l$,  it is impossible to conclude that $l$ is weakly facultative. Thus, if we consider Example~\ref{ex:ambiguity1}, then we cannot conclude $\obl(c)$.

\section{An Argumentation Semantics for Weak Permission}
\label{sec:new}

We take the notions defined in the previous section and extend it to a semantics that can deal with weak permission. The new semantics is based on \rnew{and extends} the argumentation semantics for Defeasible Logic \cite{jlc:argumentation} \rnew{with conditions} \rold{. The semantics for Defeasible Logic is then extended} to deal with \rold{the} imaginary arguments. 

The first step is to use the notion of attack between arguments to define the notions of \emph{support} and \emph{undercut}. Support means that all proper subarguments of an argument are in the set of arguments that supports it; undercut means that the set of arguments that undercuts an argument supports an argument that attacks a proper subargument of the argument. This means that the premises of the undercut argument do not hold. 

\begin{definition}[Support]
\label{def:support}
A set of arguments $S$ \emph{supports} an argument $A$ if every proper subargument of $A$ is in $S$.
\end{definition}

\begin{definition}[Undercut]
\label{def:defeat}
    A set of arguments $S$ \emph{undercuts} an argument $B$ if $S$ supports an argument $A$ that attacks a proper natural subargument of $B$. 
\end{definition}

\begin{example}
\label{ex:support-undercut}
Consider a Deontic Argumentation Theory where  $R$ contains the following rules:
\[
\begin{gathered}
    r_1\colon {} \Rightarrow a \qquad\qquad
    r_2\colon a\Rightarrow \obl(b)\\
    r_3\colon \perm_w(\neg b) \Rightarrow d \qquad
    r_4\colon {} \Rightarrow \perm(\neg b) \qquad
    r_5\colon \perm(\neg b) \Rightarrow e
\end{gathered}
\]
The set of arguments is as follows:
\begin{gather*}
    I_1\colon \perm_w(a) \qquad 
    I_2 \colon \perm_w(\neg a) \qquad 
    I_3\colon \perm_w(b) \qquad 
    I_4\colon \perm_w(\neg b) \\
    \begin{aligned}
    A_1\colon &\Rightarrow a \qquad& 
    A_2\colon A_1 &\Rightarrow \obl(b) & 
    A_3\colon A_2 &\Rightarrow \perm(b) \\
    B_1 \colon I_4 &\Rightarrow d &
    B_2\colon &\Rightarrow \perm(\neg b)\qquad& 
    B_3\colon B_2 &\Rightarrow e
    \end{aligned}
\end{gather*}
and the attack relation is as follows:
\begin{gather*}
 A_2 > I_4 \qquad A_2 > B_1 \qquad A_2 > B_2 \qquad A_2 > B_3 \\
 B_2 > A_2 \qquad B_2 > A_3 
\end{gather*}
Let us consider the set of arguments that are supported and undercut by $\emptyset$. All imaginary argument are supported by $\emptyset$, and so are arguments $A_1$ and $B_2$, because they do not have proper subarguments.  For the undercut, we have that $A_2$ is undercut by $B_2$, because $B_2$ attacks $A_2$.  

Suppose that $S=\{A_1\}$. In addition to the arguments supported by $\emptyset$, the set $S$ supports $A_2$: the only proper subargument of $A_2$ is $A_1$, which is in $S$.  Therefore, the only arguments that can be undercut are the argument attacked by $A_2$, namely, $I_4$, $B_1$, $B_2$ and $B_3$. $I_4$ and $B_2$ do not have proper subarguments, so they cannot be undercut.  $B_1$ has a proper subargument $I_4$. However, even if $A_2$ attacks $I_4$, $I_4$ is an imaginary argument, thus $S$ does not undercut $B_1$. Finally, $B_3$ has a proper subargument $B_2$, which is a natural argument and it is attacked by $A_2$, thus $S$ undercuts $B_3$. 
\end{example}

At this stage, we can define the notion of \emph{wp-acceptable} and \emph{wp-rejected} arguments. The main difference is that we have to distinguish between natural and imaginary arguments.  An imaginary argument is an argument for a weak permission, and weak permission means that it is impossible to have a valid argument for the obligation to the contrary. In other words, weak permission succeeds when we fail to derive the obligation to the contrary.  Accordingly, we must be able to tell when an argument is acceptable or when an argument is rejected.  Given that weak permission requires that we reject the obligation of the contrary, an imaginary argument is acceptable when all arguments for the prohibition are rejected.  A natural argument is acceptable when we can show that the arguments that attack it cannot fire (their premises do not hold).

\begin{definition}[wp-Acceptable]
\label{def:wp-accept}
An argument $A$ is \emph{wp-acceptable} by the sets of arguments $R$ and $S$ iff
\begin{enumerate}
    \item $A$ is an imaginary argument and $\forall B\in\Args$, $B>A$, $B$ is wp-rejected by $R$ and $S$; or
    \item $A$ is a natural argument, $S$ supports $A$, and $\forall C\in\Args$, $C>A$, either $S$ undercuts $C$ or $C\in R$. 
\end{enumerate}
\end{definition}
The next definition \rold{captures the notion of rejected argument}\rnew{specifies when an argument is rejected}. Specifically, an imaginary argument is rejected when there is an argument for the obligation to the contrary that is acceptable (thus, we did not fail to establish the obligation to the contrary). A natural argument is rejected when one of its subarguments is rejected or when there is an argument attacking it that is supported by the already accepted arguments. Thus, there is an applicable rule for the opposite of the conclusion of the (rejected) argument
\begin{definition}[wp-Rejected]
An argument $A$ is \emph{wp-rejected} by the sets of arguments $R$ and $S$ iff
\begin{enumerate}
\item $A$ is an imaginary argument and $\exists B\in\Args$, $B>A$, $B\in S$; or
\item $A$ is a natural argument and 
    \begin{enumerate}
        \item $\exists B\in\sub(A)$, $B\neq A$, and $B\in R$; or
        \item $\exists B\in\Args$, $B>A$, and $S$ supports $B$.
    \end{enumerate}
\end{enumerate}
\end{definition}
The wp-semantics (or wp-extension) is then defined by the following fixed-point construction, where we iteratively build the sets of wp-acceptable and wp-rejected arguments. Notice that since we have to capture both what arguments are acceptable and what arguments are rejected, the extension is given by the pair of such sets of arguments.
\begin{definition}[wp-Semantics]
The \emph{wp-extension} of a Deontic Argumentation Theory $T$ is the pair
\[
(\mathit{JArgs},\mathit{RArgs})
\]
such that 
\[
    \mathit{JArgs}=\bigcup_{i=1}^{\infty}J^T_i
    \qquad\qquad
    \mathit{RArgs}=\bigcup_{i=1}^{\infty}R^T_i
\]
where
\begin{itemize}
    \item $J^T_0=\emptyset$; $R^T_0=\emptyset$;
    \item $J^T_{n+1}=\{ A\in\Args: A \text{ is wp-acceptable by } R^T_n \text{ and } J^T_n\}$;
    \item $R^T_{n+1}= \{A\in\Args: A \text{ is wp-rejected by } R^T_n \text{ and } J^T_n \}$.
\end{itemize}
\end{definition}
Whenever clear from the context, we will drop the references to the sets $R$ and $S$ when speaking of wp-accepted and wp-rejected arguments.

\begin{example}
\label{ex:wp-extensive}
Let us consider again the Deontic Argumentation Theory of Example~\ref{ex:extensive}. The arguments and the instances of the attack relation are the same for the grounded semantics and the wp-semantics.

Let us go step-by-step through the construction of the wp-extension.  The first step is to determine the arguments that are wp-acceptable and wp-rejected by $J^T_0$ and $R^T_0$, namely the empty set. To this end, we start by listing what arguments are supported and undercut by $J^T_0$ and $R^T_0$.
It is easy to verify that the only arguments supported by $J^T_0$ and $R^T_0$ are
\[
 A_0, A_1, A_2, A_4.
\]
These are the arguments that do not have any proper subarguments (we ignored all imaginary arguments since they are not involved in any attack, and the condition to ). The next step is to determine the arguments that are undercut, namely:
\[
 A_3, A_5, A_6, A_7, A_{10}.
\]
$A_6$ is undercut because it is attacked by $A_1$; the arguments $A_3,A_5,A_7$ are undercut because they are attacked by $A_2$, and, finally, $A_{10}$ because it is attacked by $A_4$. 

Then, we can compute $J^T_1$ and $R^T_1$:
\begin{align*}
    J^T_1 = \{A_0,A_4\}  &&
    R^T_1 = \{A_1,A_2,A_3,A_5,A_6,A_7,A_{10}\}
\end{align*}
Argument $A_0$ is trivially wp-acceptable because no argument attacks it, and it does not have any proper subarguments. $A_4$ is attacked by $A_3$ and $A_7$, but these two arguments are undercut by $J^T_0$. 
 
Arguments $A_1$, $A_5$, $A_7$, and $A_{10}$ are wp-rejected because they are attacked by $A_2$, which is supported by $J^T_0$. Arguments $A_2$ and $A_6$ are wp-rejected because they are attacked by $A_1$, which is supported by $J^T_0$. Argument $A_3$ is wp-rejected because it is attacked by $A_4$, which is supported by $J^T_0$.       

We can now move to the next step, where we compute $J^T_2$ and $R^T_2$.  In addition to the arguments supported by $J^T_0$, $J^T_1$ supports $A_8$ ($A_4\in J^T_1$). 
\begin{align*}
    J^T_2 = J^T_1 \cup \{I_1,I_2,I_4,A_8\} &&
    R^T_2 = R^T_1 \cup \{I_3\}
\end{align*}
The imaginary arguments $I_1$, $I_2$ and $I_4$ are wp-acceptable because they are attacked respectively by $A_2$, $A_1$ and $A_3$, and these arguments are wp-rejected by $R^T_1$. Argument $A_8$ is wp-acceptable because it is now supported by $J^T_1$ and it is attacked by $A_3$ and $A_7$. $A_3$ and $A_7$ are undercut by $J^T_1$. $I_3$ is wp-rejected because it is attacked by $A_4$, which is wp-acceptable by $J^T_1$.

For the next and final step, we compute $J^T_3$ and $R^T_3$. This time, $J^T_2$ adds support for argument $A_9$ (because $I_1\in J^T_2$), and then $A_9$ is wp-acceptable, because it is attacked by $A_2$, but $A_1$ attacks $A_2$ and it is supported by $J^T_2$.
\end{example}

\begin{theorem}
\label{thm:wep-semantics}
    For any Deontic Argumentation Theory $T=(F,R)$ the wp-extension is unique.
\end{theorem}
\begin{proof}
The proof that the wp-extension is unique is by induction on the steps of the construction of the wp-extension, that if an argument is wp-\rchange{justified}{acceptable}/wp-rejected by $R^T_n$ and $J^T_n$, then the argument is wp-\rchange{justified}{acceptable}/wp-rejected by $R^T_{n+1}$ and $J^T_{n+1}$. The induction proves that the extension of the wp-extension is monotonically increasing, and then we can apply the Knaster-Tarski Theorem to conclude that the wp-extension is the unique least/greatest fixed-point. 

Let us start by the case when an argument $A\in J^T_1$. We have two sub-cases.

1) $A$ is an imaginary argument. Thus, it does not have any proper subargument (so no argument can be wp-rejected), and every argument attacking it is rejected by $\emptyset$ and $\emptyset$ ($R^T_0$ and $J^T_0$).  Let $B$ be an argument attacking $A$. Since $A$ is wp-acceptable, then $B$ is wp-rejected; therefore there is an argument $C$, $C>B$ and 
$C$ is supported by $\emptyset$. This means that $C$ does not have any proper subarguments, which in turn means that any $J^T_i$ supports $C$ rejecting $B$ in any iteration of the construction. Hence, $A\in J^T_1$.

2) $A$ is a natural argument. By construction $A$ is wp-acceptable by $R^T_0$ and $J^T_0$; by definition, $A$ is supported by $J^T_0$, namely $\emptyset$, thus, again, $A$ does not have any proper subarguments. Accordingly, $A$ is supported by any $J^T_i$. The arguments attacking it are undercut by $J^T_0$. This means that for every argument $B$ attacking $A$, there is an argument $C$ supported by $J^T_0$. But as we have just seen for $A$, the argument $C$ is supported by any $J^T_i$. So, $A\in J^T_1$.

We can now move to the case of a wp-rejected argument. Thus, $A\in R^T_1$. Again, we have two sub-cases.

1) $A$ is an imaginary argument. $A$ is wp-rejected by $R^T_0$ and $J^T_0$, if there is an argument attacking $A$ that is supported by $J^T_0$. We have already seen that an argument supported by $J^T_0$ is supported by any $J^T_i$. Accordingly, $A$ is wp-rejected at any step of the construction of the wp-extension; hence $A\in R^T_1$.

2) $A$ is a natural argument. Given that $R^T_0=\emptyset$, $A$ cannot be in $R^T_0$ because one of its proper subarguments is in $R^T_0$. Thus, there is an argument $B$ attacking in that is supported by $J^T_0$. We can argue, as in the other case, that $B$ is supported by any $J^T_i$, and thus, $A\in R^T_1$. 

\smallskip

For the inductive step, as usual, we assume that the property holds for arguments in $J^T_n$ and $R^T_n$. Hence, for the justified arguments we can conclude that if an argument is wp-acceptable for some set of arguments $J^T_k$ ($k<n$), then the argument is supported by $J^T_n$; at this stage we can replicate the reasoning of the inductive base to conclude that if an argument is wp-acceptable by $R^T_n$ and $J^T_n$, then it is wp-acceptable by $R^T_{n+1}$ and $J^T_{n+1}$.

For the case when an argument $A$ is in $R^T_{n+1}$ in addition to the cases discussed in the inductive base that carry over by the inductive hypothesis, we have to examine the situation that there a proper subargument $A'$ of $A$ is wp-rejected by $R^T_n$ and $J^T_n$. This means that $A'\in R^T_n$, and that argument has been added at a step before $n$. Thus, we can appeal to the inductive hypothesis to conclude that $A'$ would be in all $R^T_m$, $m\geq n$. Thus, $A\in R^T_{n+1}$.

\smallskip

We have just proved that the construction of the wp-extension is monotonically increasing. Then, by the Knaster-Tarski Theorem, the construction has a least fixed point, and the least fixed point is unique. 
\end{proof}

The next step is to show that the wp-semantics is correct. For correctness, we have to consider two properties. We have to show that the wp-extension is conflict-free, thus the set of wp-acceptable arguments does not contain arguments for and against a given conclusion and that no argument is both wp-justified and wp-rejected.
    
\begin{theorem}
\label{thm:wp-correctness}
    For any Deontic Argumentation Theory $T=(F,R)$:
    \begin{itemize}
        \item $\Jargs\cap\Rargs=\emptyset$ (no argument is both wp-\rchange{justified}{acceptable} and wp-rejected);
        \item the wp-extension is conflict-free.
    \end{itemize}
\end{theorem}
\begin{proof}
We start by proving that the intersection of the wp-acceptable argument and the wp-rejected sets of arguments is empty. The proof is by contradiction and induction.
Assume that there is an argument $A$ such that $A\in\Jargs$ and $A\in\Rargs$. This means that there is some $n$, $A\in J^T_n$ and $A\in R^T_n$. We prove by induction that this is impossible. 

For the inductive base, suppose that $A\in J^T_0$ and $A\in R^T_0$.  We have two cases: $A$ is an imaginary argument or $A$ is a natural argument. 

$A$ is an imaginary argument. $A\in R^T_1$ means that there is an argument $B$, such that $B>A$ and $B\in J^T_0$. But $J^T_0=\emptyset$, so there is no such argument $B$. Hence, $A\notin R^T_1$. Thus, $A$ cannot be in the intersection of $J^T_1$ and $R^T_1$. 

$A$ is a natural argument. $A\in J^T_1$ means that $J^T_0=\emptyset$ supports $A$, thus $\sub(A)=\{A\}$ and $A$ does not have any proper subarguments. Moreover, $J^T_0$ undercuts all arguments attacking $A$.  Suppose $B>A$, then $J^T_0$ undercuts $B$, thus $J^T_0$ supports an argument $C$ that attacks a proper subargument of $D$ of $B$, $D\in\sub(B),D\neq B$.
$A\in R^T_1$ means that there is an argument $B$, such that $B>A$ and $J^T_0=\emptyset$ supports $B$. Thus, $\sub(B)=\{B\}$, and we get a contradiction since $C$ cannot attack a proper subargument $D$ of $B$.

For the inductive step, we assume that the property holds for $n$  ($J^T_n\cap R^T_n=\emptyset$) and we show that it holds for $n+1$. Assume that $A\in J^T_n$ and $A\in R^T_n$. Again, we have two cases: $A$ is an imaginary argument or $A$ is a natural argument.

$A$ is an imaginary argument. $A\in J^T_{n+1}$ means that every argument attacking $A$ is wp-rejected by $R^T_n$ and $J^T_n$. Thus, any argument $B$, such that $B>A$, is in $R^T_n$.  $A\in R^T_{n+1}$ means that there is an argument $B$, such that $B>A$ and $C\in J^T_n$. Hence,  $B\in J^T_n\cap R^T_n$, against the inductive hypothesis. 

$A$ is a natural argument. $A\in J^T_{n+1}$ means that $J^T_n$ supports $A$, 
and every argument $B$ attacking $A$ is undercut by $J^T_n$. Thus, for every argument $A'\in\sub(A), A'\neq A$, $A'$ is in $J^T_n$.  Moreover, for an argument $B$ such that $B>A$, $J^T_n$ supports an  argument $C$ that attacks a proper subargument of $B'$ of $B$ ($B'\in\sub(B), B'\neq B$). $A'\in\sub(A), A'\neq A$ such that $A'\in R^T_n$ (and this is against the inductive hypothesis) or (2) there is an argument $B$ such that $B>A$ and $J^T_n$ supports $B$. Therefore, $\forall B'\in\sub(A), B'\neq A$, $B'\in J^T_n$. But, $B'$ is undercut by $J^T_n$, thus there is an argument $C$ supported by $J^T_n$ that attacks $B'$. Thus, $B'\in R^T_n$ against the inductive hypothesis.

\addvspace{9pt}

For the wp-semantics, conflict-free means that we do not have pairs of arguments $A$ and $B$ such that $A>B$ and $A,B\in\Jargs$. 

Let us suppose that an extension is not conflict-free; this means that there is some $n$ such that there are two arguments $A$ and $B$ such that $A,B\in J^T_{n+1}$, and $A>B$. By construction of the attack relation, we have that $A$ is a natural argument. Thus, $A\in J^T_{n+1}$ means that $J^T_n$ supports $A$, and every argument $C$ attacking $A$ is undercut by $J^T_n$.  

$B$ is either an imaginary argument or a natural argument. If $B$ is an imaginary argument, $B\in J^T_{n+1}$ means that every argument attacking $B$ is wp-rejected by $R^T_n$ and $J^T_n$. But, $A$ attacks $B$, thus $A$ is wp-rejected by $R^T_n$ and $J^T_n$, so $A\in R^T_{n+1}$; but we just proved that this is impossible, no argument is both wp-\rchange{justified}{acceptable} and wp-rejected.

If $B$ is a natural argument, $A$ attacks $B$, and $J^T_n$ supports $A$, thus, $B\in R^T_{n+1}$. But, as we have just seen, this is impossible.
\end{proof}    

As we did for the grounded semantics, we can extend the notion of wp-justified conclusions to literals and deontic literals. A literal is wp-justified if it is the conclusion of a wp-justified argument.

\begin{definition}[wp-Justified]
\label{def:wp-justified}
    Given a Defeasible Argumentation Theory $T$,
    an argument $A$ is \emph{justified} under the wp-semantics (or wp-justified) iff $A\in\Jargs$. 
    
    A literal or a deontic literal is \emph{justified} under the wp-semantics (or wp-justified) iff it is the conclusion of a justified argument under the wp-semantics.  
\end{definition}
    
\begin{example}
    \label{ex:wp-justified}
    When we consider the Deontic Argumentation Theory of Example~\ref{ex:extensive}, and the construction of the extension in Example~\ref{ex:wp-extensive}, we can see that the arguments $A_0$, $A_4$, $A_8$, $A_9$, $I_1$, $I_2$ and $I_4$ are wp-justified. Thus, the wp-justified conclusions are 
    \[
    a, \obl(\neg q), \perm(\neg q), s, \perm_w(p), \perm_w(\neg p), \perm_w(\neg q)
    \]
\end{example}    

The final step is to show that weak permission is supported by the wp-semantics. This means that we can infer that a weak permission is in force even when a conflict between two obligations exists. Accordingly, we refine the definition of a conflictual theory to be more aligned with the wp-semantics. 

\begin{definition}
\label{def:wp-conflictual}
A Deontic Argumentation Theory is \emph{wp-conflictual} it contains a pair of rules 
\begin{equation*}
a_0,\dots,a_n\Rightarrow\obl(l) \qquad\qquad
b_0,\dots,b_m\Rightarrow\obl(\neg l)
\end{equation*}
such that the arguments $A_i$ ($0\leq i\leq n$) and $B_j$ ($0\leq j\leq m$) have conclusions $C(A_i)=a_i$ and $C(B_j)=b_j$, and the $A_i$ and $B_j$ are wp-justified. We also say that $l$ is a wp-conflictual literal. 
\end{definition}

\begin{theorem}
\label{lem:weak-permission}
    Let $T$ be a wp-conflictual theory and $l$ a wp-conflictual literal such that the proper subarguments of the arguments for $\obl(l)$ and $\obl(\neg l)$ are wp-justified. Then $\perm_w(l)$ and $\perm_w(\neg l)$ are wp-justified.
\end{theorem}
\begin{proof}
\rnew{By definition, there are two rules
\begin{align*} 
b_1,\dots,b_n \Rightarrow\obl(l) &&
c_1,\dots,c_m \Rightarrow\obl(\neg l)
\end{align*}
with arguments
\begin{align*}
A_1\colon B_1,\dots,B_n &\Rightarrow \obl(l) &
A_2\colon C_1,\dots,C_m &\Rightarrow \obl(\neg l)
\end{align*}
such that the (sub-)arguments $B_j$ ($1\leq j\leq n$) and $C_k$ ($1\leq k\leq m$) are wp-justified. Given that the arguments $B_j$ and $C_k$ are wp-justified,} can assume, without any loss of generality, that the arguments $B_j$ and $C_k$ have no attacks.  Moreover, Hence, the set of attacks for the theory is 
\begin{align*}
 A_1 > A_2 && A_1 > \perm_w(\neg l)\\
 A_2 > A_1 && A_2 > \perm_w(l)
\end{align*}
By definition all subarguments of $A_1$ and $A_2$ are wp-justified. $A_1$ is the only argument attacking $\perm_w(\neg l)$, $A_2$ attacks $A_1$ and $A_2$ is supported by the wp-extension. Hence, $A_1$ is wp-rejected. $A_1$ is the only argument attacking $\perm_w(\neg l)$. Hence, $\perm_w(\neg l)$ is a justified argument. We can repeat the same argument for $\perm_w(l)$. 
\end{proof}

\rnew{Before we move to the next set of results we show how the wp-semantics works on the Sea Watch 3 scenario.}

\rnew{
\begin{example}\label{ex:sea-watch-wp}
Let us consider again the Deontic Argumentation Theory modelling the Sea Watch 3 scenario (Example~\ref{ex:sea-watch-arg}), but this time we analyse it from the perspective of wp-semantics. The arguments and the attack relation are the same as under the grounded semantics.  

Given that there are no attacks and no proper subarguments for the arguments $A_1$, $A_2$, $A_3$, $A_4$, $A_5$, and $A_6$, these arguments are supported by $\emptyset$ and are not undercut by $\emptyset$. Hence, these arguments are in $J^T_1$.

Given that $\sub(A_7)=\{A_7, A_1, A_2\}$ and that $A_1$ and $A_2$ are in $J^T_1$, $A_7$ is supported by $J^T_1$. Similarly, since $\sub(A_8)=\{A_8, A_3, A_4, A_5\}$ and $A_3$, $A_4$, and $A_5$ are in $J^T_1$, $A_8$ is supported by $J^T_1$.

$A_7$ is attacked by $A_8$, which is supported by $J^T_1$; thus, $A_7$ is wp-rejected by $R^T_1$. Likewise, $A_8$ is attacked by $A_7$, which is supported by $J^T_1$; thus, $A_8$ is wp-rejected by $R^T_1$.

This means that arguments $I_1$ and $I_2$ are wp-justified by $J^T_1$ and $R^T_1$, since they are attacked by $A_8$ and $A_7$, respectively, and both of those arguments are wp-rejected by $R^T_1$. Hence, $I_1$ and $I_2$ are in $J^T_2$. Argument $A_9$ is then supported by $J^T_2$, given that $A_6$ and $I_2$ are in $J^T_2$, and it is attacked by $A_8$, which is wp-rejected by $R^T_1$. Thus, $A_9$ is wp-justified by $J^T_2$ and $R^T_2$.
\end{example}
}

Contrary to the grounded and well-founded semantics, weakly facultative conclusions are wp-justified.
\begin{corollary}
    Let $D=(F,R)$ be a wp-conflictual theory and $l$ a wp-conflictual literal, then $l$ is a weakly facultative conclusion.
\end{corollary}
Let us examine the scenario of Example~\ref{ex:ambiguity1} under the wp-semantics, where the permission in $n_3$ are understood as a weak permission. Namely, 
\[
 n_3\colon \perm_w(a),\perm_w(\neg a) \Rightarrow \obl(c)
\]
The (relevant) imaginary arguments are:
\[
I_1\colon \perm_w(a) \qquad
I_2\colon \perm_w(\neg a) \qquad
I_3\colon \perm_w(c) \qquad
I_4\colon \perm_w(\neg c)
\]
and the natural arguments are:
\begin{align*}
A_1\colon F_1,\dots,F_n &\Rightarrow \obl(a) &
A_2\colon G_1,\dots,G_m &\Rightarrow \obl(\neg a)\\
A_3\colon I_1,I_2 &\Rightarrow \obl(c) &
A_4\colon H_1,\dots,H_k &\Rightarrow \obl(\neg c)
\end{align*}
The attack relation includes the following instances:
\begin{align*}
\qquad && A_1 &> I_2 & A_1 &> A_2 & A_1 &> A_3 &&\qquad\\
\qquad && A_2 &> I_1 & A_2 &> A_1 & A_2 &> A_3 &&\qquad\\
\qquad && A_3 &> A_4 & A_3 &> I_4 & A_4 &> I_1 &&\qquad
\end{align*}
Let us assume that for some $n$, $\{F_1,\dots,F_n,G_1,\dots,G_m,H_1,\dots, H_k\}\in J^T_n$. This means that $A_1,A_2,A_4$ are supported by $J^T_n$.
Then, $A_1$ is attacked by $A_2$, which is supported by $J^T_n$. Thus, $A_1$ is wp-rejected by $J^T_n$ and $R^T_n$, and so $A_1\in R^T_{n+1}$. For a similar reason $A_2\in R^T_{n+1}$. Now $I_1$ is attacked by $A_2$, and $I_2$ by $A_1$, but $A_1$ and $A_2$ are wp-rejected by $J^T_{n+1}$ and $R^T_{n+1}$. Thus, $I_1,I_2$ are wp-justified by $J^T_{n+1}$ and $R^T_{n+1}$, and so $I_1,I_2\in J^T_{n+2}$.   At this stage, $A_3$ is supported by $J^T_{n+2}$ and has no attacks. Hence, $A_3$ is wp-justified by $J^T_{n+2}$ and $R^T_{n+2}$, and so $A_3\in J^T_{n+3}$. Finally, $A_4$ is attacked by $A_3$, which is supported by $J^T_{n+2}$, thus $A_4$ is wp-rejected by $J^T_{n+3}$ and $R^T_{n+3}$, and so $A_4\in R^T_{n+4}$. This means that $\perm_w(a)$, $\perm_w(\neg a)$ and $\obl(c)$ are wp-justified.

To conclude we show that the wp-semantics is an extension of the grounded semantics. To this end we need the following lemma.
\begin{lemma}\label{lem:subarguments}
    All subarguments of a wp-justified argument are wp-justified.
\end{lemma}
\begin{proof}
It is an immediate consequence of Definition~\ref{def:wp-accept}.
\end{proof}

\begin{theorem}
\label{thm:inclusion}
    Given a deontic argumentation theory, every argument justified under grounded semantics is also justified under the wp-semantics.
\end{theorem}
\begin{proof}
\rnew{
To prove this result, we use Dung's fixed-point construction of grounded semantics \cite{Dung1995} and proceed by induction on its stages. We show that if an argument is justified at stage $n$ in the construction of the grounded extension, then it is wp-justified.

The fixed-point construction of the grounded extension is defined as follows~\cite{Dung1995}:
\begin{itemize}
    \item $S_0=\emptyset$;
    \item $S_{n+1} = \{A\in\Args: A \text{ is acceptable with respect to } S_n\}$.
\end{itemize}
For the inductive base, we have $S_0=\emptyset$, and thus we consider the arguments that are acceptable with respect to the empty set. An argument $A$ is acceptable with respect to the empty set if every argument attacking $A$ is attacked by some argument in the empty set. This means that no argument attacks $A$. We then have two cases: $A$ is either an imaginary argument or a natural argument.

Suppose first that $A$ is an imaginary argument. Since $A$ is acceptable with respect to the empty set, no argument attacks $A$. Therefore, $A$ is wp-acceptable with respect to the empty set and the empty set, and thus $A\in J^T_1$.

Now suppose that $A$ is a natural argument. We must show that $A$ is supported by some $J^T_n$. To this end, consider the set of subarguments of $A$. By definition, an argument is rooted in the set of facts, and every fact is in $J^T_1$. Arguments whose only proper subarguments are facts are therefore supported by $J^T_1$ and belong to $J^T_2$. Repeating the same reasoning for all proper subarguments of $A$, we conclude that $A$ is supported by some $J^T_m$, and thus $A$ is wp-acceptable with respect to $J^T_{m+1}$. Hence, $A\in\Jargs$.

For the inductive step, assume that the property holds for $n$. Thus, every argument that is acceptable with respect to $S_n$ is wp-justified. We must show that every argument that is acceptable with respect to $S_{n+1}$ is wp-justified. Let $A$ be an argument that is acceptable with respect to $S_{n+1}$. This means that every argument $C$ attacking $A$ is attacked by some argument $B\in S_n$. By the inductive hypothesis, $B$ is wp-justified. Thus, there is some $J^T_m$ such that $B\in J^T_m$. By Lemma~\ref{lem:subarguments}, all subarguments of $B$ are in $J^T_m$. Thus, $B$ is supported by $J^T_m$. It follows that $C$ is attacked by an argument supported by $J^T_m$. Hence, $C$ is wp-rejected with respect to $R^T_m$ and $J^T_m$, and thus $A$ is wp-acceptable with respect to $R^T_{m+1}$. Therefore, $A\in\Jargs$. }   
\end{proof}    

\section{Discussion}
\label{sec:discussion}

The novel semantics was developed to address an issue with other argumentation semantics, namely the inability to deal with weak permission in the presence of conflicts.   A classical example of this issue is the case of 
\emph{moral dilemmas} where two obligations conflict. Thus, we have a pair of norms $a\Rightarrow \obl(p)$ and $b\Rightarrow \obl(\neg p)$.
 
According to our semantics, the two norms \emph{cancel each other} in the sense that we do not allow the two norms to be justified together, and we prevent the derivation of the two obligations.  Here, to understand whether preventing the two obligations is meaningful, we must consider the sources for the two norms. 
If the norms originate from different legal systems, they do not cancel each other, and a conflict does not exist. This means that it is impossible to comply with both norms. But, formally, we should have $\obl_1(p)$ and $\obl_2(\neg p)$. So, while it is impossible to comply with both of them, then two legal systems are not (individually) inconsistent.
 
If the two norms are from the same legal system, then we have an inconsistency. Here, the inconsistency can be understood again in the impossibility of complying with all justified obligations. Assuming that violations lead to sanctions, if the two obligations are justified, a subject of the legal system will be sanctioned no matter what, against a well-established principle of legal theory, namely that a subject should not be sanctioned for complying with the law.  

If there are mechanisms to resolve the conflict, one of the two norms will prevail over the other, and only one obligation will be justified, and no inconsistency will arise.  However, if there are no easily identifiable mechanisms to resolve the conflict, then if we consider only one of the two as justified, then we are assuming that the legal system had a mechanism to solve the conflict, but this is contrary to what we assumed, namely that the two norms are in conflict and no mechanism was available. 

Suppose that the proposed resolution happens in the course of a legal proceeding.  However, there is no way in which the parties involved in the legal proceedings know in advance what the resolution will be. Thus, asserting that one of the obligations is justified would amount to an \emph{ad hoc} adjudication, and possibly to a retroactive change to the legal system (the resolution was not available at the time the parties were acting based on the two conflicting norms). Ad hoc and retroactive applications are considered violations of the so-called principles of legality for legal systems and should be avoided \cite{fuller1969morality}.  

Accordingly, the only remaining option is that none of the two obligations is justified. Thus, we fail to assert that any of the two obligations is in force. But the failure to obtain a (justified) obligation corresponds to the assertion of a weak permission. Thus, we can conclude that in the presence of conflicting obligations, we can assert a weak permission for both obligations.

In this paper we distinguished between weak permission ($\perm_w$) and strong permission ($\perm$). Typically in legal provisions there is not a distinction between weak and strong permission, and provisions will simply use the term permission, and only the usage (and the context) determines whether the permission is weak or strong (as we did in Section~\ref{sec:intuitions}). Provisions mandating that something is permitted are interpreted as strong permission. When a provision uses a permission as one of the conditions of applicability of the provision, then the permission is interpreted as the disjunction of strong and weak permission.  It is possible to have provisions that use strong permission (for example, using terms like authorised) or weak permission (for example, using expressions like “provided that $p$ is not forbidden”\footnote{Alchourron \cite{alchourron-bulygin:1969} suggested the expression “it is false that $x$ has issued a norm to the effect that $p$ is forbidden” (where $x$ refers to an underlying normative system) for weak permission.}).

\section{Related Work}
\label{sec:related}

Formal argumentation has been investigated for a long time in the field of Artificial Intelligence and Law (for a comprehensive \rold{but somewhat}\rnew{though a bit} outdated survey, see \cite{PrakkenS15}). However, the issue of incorporating deontic reasoning in argumentation frameworks has been neglected. Moreover, the issue of weak permission received almost no attention. \rnew{Notable exceptions are the works by Beirlaen, Heyninck, and Straßer \cite{10.1093/logcom/exy005} and the work on Defeasible Deontic Logic \cite{jpl:permission}}.

The work by van der Torre and Villata \cite{TorreVillata} extends ASPIC$^+$ \cite{ModgilPrakken2014} with deontic operators (obligation) and analyses the framework in terms of Input/Output logic. However, it does not consider weak permission. 

The scope of \cite{daCosta} \rold{was the use of}\rnew{is} formal argumentation for handling norms in multi-agent systems, presenting methods for representing and interpreting hierarchical normative systems, and addressing open texture in legal interpretation using fuzzy logic. The work does not focus on deontic reasoning or weak permissions: deontic assertions are simply treated as conclusions in an ASPIC$^+$ setting.

The aim of \cite{PigozziTorre} is to propose an argumentation framework to address different types of normative conflicts. Arguments are built by triples $(b,i,d)$ where $b$ is a formula for the brute facts of the arguments, $i$ represents the institutional facts, and finally $d$ is the deontic conclusion of the argument, which is either an obligation or a permission. While the work considers various notions of conflicts, it does not consider weak permission. Moreover, the argumentation framework adopts Dung's semantics, including grounded semantics. Thus, it cannot deal with weak permission when unresolved conflicts exist.

Riveret, Rotolo and Sartor \cite{RiveretRotoloSartor} presented a modular rule-based argumentation system designed to represent and reason about conditional norms, including obligations, prohibitions, and weak and strong permissions. However, the work treated such notions as labels of literals without exploring their argumentative nature. Similar considerations apply to \cite{RiverertOrenSartor}, in which deontic argumentation and probabilistic argumentation were merged. 

The work by Straßer, Arieli and Berkel \cite{StrasserArieli,ArieliBerkelStrasser} focuses on the relationships between formal argumentation and sequent style and Input/Output inference systems. However, in general, they do not consider permission (and weak permission) \cite{ArieliBerkelStrasser} or Standard Deontic Logic \cite{StrasserArieli}. While \cite{ArieliBerkelStrasser} restricts its \rold{attention}\rnew{focus} to grounded semantics, \cite{Strasser:comma22,Strasser:comma24} consider also other semantics (preferred and stable) but no permission. \rnew{\cite{10.1093/logcom/exy005} includes a discussion of weak permission, and proposes including arguments for weak permission in a manner essentially identical to our approach but framed in the context of the grounded semantics. However, as we have shown in Theorem~\ref{thm:impossibility} and the results of \cite{weakpermission}, the grounded semantics is not able to characterise weak permission, and thus the approaches above are not able to fully capture the notion of weak permission. }

The proposal in \cite{YuLu} builds an argumentation framework in ASPIC$^+$ style for deontic reasoning based on Standard Deontic Logic extended with a preference relation.  The approach considers permission as the dual of obligation, and then evaluates arguments based on the complete semantics. Notice that in case of a conflict, the complete semantics will provide two extensions, and the proof of Theorem \ref{thm:impossibility} shows the two extensions contain opposite obligations, and weak permission is not sceptically justified.   

The approach closest to our work is \rold{the one advanced by} Defeasible Deontic Logic (DDL) \cite{jpl:permission}. DDL can deal with conflicts and weak permission; DDL offers a constructive proof-theoretic approach to deontic reasoning, though its proof theory has a strong argumentation flavour. However, no argumentation semantics has been advanced for DDL. The wp-semantics is based on the argumentation semantics for Defeasible Logic \cite{jlc:argumentation}.  
Given that wp-semantics extends the semantics proposed in \cite{jlc:argumentation} for defeasible logic to account for deontic operators, including weak permission, we expect wp-semantics to characterise DDL. A preliminary analysis in argumentation was developed in \cite{GovernatoriRotoloRiveret}. In this paper, an initial treatment of weak permission was developed by introducing the concept of argument agglomeration set, with respect to which an argument for a prohibition is rejected, thus proving the corresponding weak permission. However, no systematic study on argumentation semantics was offered, thereby confining the study to DDL.

Grounded semantics and the semantics proposed in \cite{jlc:argumentation} correspond (respectively) to ambiguity propagation and ambiguity blocking. \cite{icail2011carneades,icail23} point out that legal and deontic reasoning support ambiguity blocking and ambiguity propagation. In general, the two approaches are incompatible.  \cite{icail23} proposed an extension of DDL, DDL$^+$ that is a conservative extension of the variants.  DDL$^+$ has been defined proof-theoretically. However, DDL$^+$ is a conservative extension of DDL. Accordingly, we plan to investigate an argumentation framework that allows us to integrate the two competing semantics. 

While the wp-semantics covers the case of weak permission, it fails to explain the deontic operators (apart from weak permission being the failure to conclude the prohibition of the opposite). To fill this gap, we plan to extend the approach we proposed in \cite{jelia23} to account for the meaning of the deontic operators in a possible world (neighbourhood) semantics for Defeasible Deontic Logic \cite{deon2012}.

\section{Summary}
\label{sec:conclusion}

In this paper, we have proposed a new semantics for deontic argumentation that can deal with weak permission. The semantics is based on the argumentation semantics for Defeasible Logic. We have shown that the semantics is unique, conflict-free, and that no argument is both justified and rejected. We have also shown that the semantics is able to characterise weak permission when there are unresolved conflicts, thereby addressing the issue with grounded semantics.

\bibliographystyle{plainnat}
\bibliography{deon}

\end{document}